\pdfoutput=1

\documentclass[11pt]{article}

\usepackage{emnlp2021}

\usepackage{times}
\usepackage{latexsym}

\usepackage[T1]{fontenc}

\usepackage[utf8]{inputenc}

\usepackage{microtype}

\usepackage{graphicx}
\usepackage[utf8]{inputenc}
\usepackage{amsmath,amssymb,amsfonts,bbm}
\usepackage[mathcal]{euscript}
\usepackage[ruled,vlined]{algorithm2e}
\usepackage{algpseudocode}
\usepackage{tabularx}
\usepackage{multirow}

\newcommand{\eg}{\textit{e.g.\ }}

\newcommand{\aka}{\textit{a.k.a.\ }}

\newcommand{\covidi}{COVID-I\xspace}
\newcommand{\covidii}{COVID-II\xspace}

\title{Self-Supervised Detection of Contextual Synonyms in a Multi-Class Setting: Phenotype Annotation Use Case}

\author{Jingqing Zhang$^{1,2}$, Luis Bolanos$^{2}$, Tong Li$^{2}$, Ashwani Tanwar$^{2}$, Guilherme Freire$^{2}$ \\ 
\textbf{Xian Yang$^{3}$, Julia Ive$^{1}$, Vibhor Gupta$^{2}$, Yike Guo$^{1,2,3}$} \\
$^{1}$Data Science Institute, Imperial College London, UK \\
$^{2}$Pangaea Data Limited, UK, USA \\
$^{3}$Hong Kong Baptist University, Hong Kong SAR, China\\
\texttt{{jzhang,lbolanos,tli,atanwar,gfreire,vgupta}@pangaeadata.ai} \\
\texttt{{j.ive,y.guo}@imperial.ac.uk ~xianyang@comp.hkbu.edu.hk}} 
\date{}

\begin{document}
\maketitle
\begin{abstract}


Contextualised word embeddings is a powerful tool to detect contextual synonyms. However, most of the current state-of-the-art (SOTA) deep learning concept extraction methods remain supervised and underexploit the potential of the context. In this paper, we propose a self-supervised pre-training approach which is able to detect contextual synonyms of concepts being training on the data created by shallow matching. We apply our methodology in the sparse multi-class setting (over 15,000 concepts) to extract phenotype information from electronic health records. We further investigate data augmentation techniques to address the problem of the class sparsity. Our approach achieves a new SOTA for the unsupervised phenotype concept annotation on clinical text on F1 and Recall outperforming the previous SOTA with a gain of up to 4.5 and 4.0 absolute points, respectively. After fine-tuning with as little as 20\% of the labelled data, we also outperform BioBERT and ClinicalBERT. The extrinsic evaluation on three ICU benchmarks also shows the benefit of using the phenotypes annotated by our model as features.

\end{abstract}

\section{Introduction}
\label{sec:intro}

Supervised fine-tuning on the top of the BERT-based models has recently become the standard approach in Information Extraction delivering state-of-the-art (SOTA) results across different tasks~\cite{devlin-etal-2019-bert}. The dependence of these models on the availability of the costly human-annotations remains a serious obstacle towards a large scale deployment of such models. This problem is especially actual in the clinical domain with limited availability of experts.

In the self-supervised setting, automatic annotations are cheap to produce. Some rule-based automatic labelers, such as CheXpert \cite{irvin2019chexpert}, which is built on NegBio \cite{peng2018negbio}, are often used to create training data for supervised BERT-based models (e.g., automatic annotators of radiology reports~\cite{smit-etal-2020-combining}). Those models usually generalise over a small set of classes (under 20).

Other automatic labelers exploit ontologies. For example, the UMLS (Unified Medical Language System) ontology~\cite{bodenreider2004unified} is almost predominantly used to match linguistics patterns in clinical text to medical concepts (e.g., using the MetaMap tool ~\cite{aronson2006metamap}). Due to the complexity of the Information Extraction task in this challenging setting (sparse multi-class), the approaches that use the data annotated  (e.g., ~\cite{arbabi2019ncr,kraljevic2019medcat,Tiwari2020}) mostly rely on non-contextualised embeddings focusing on the detection precision. However, especially for clinical text, which is noisier and exhibits a variety of clinical expressions requiring disambiguation, relying on the context is essential. We argue that recall is very important, especially when automatic annotation results are used further in the downstream tasks.

In this work, we propose a self-supervised approach for sparse multi-class classification that fully relies on the context to detect contextual synonyms of medical concepts in clinical text. To be more precise, our model is based on the ClinicalBERT~\citep{alsentzer-etal-2019-publicly} model which was pre-trained on the biomedical and clinical corpora that are widely used, producing state-of-the-art results in a range of supervised biomedical tasks, e.g. named entity recognition, relation extraction and question answering~\citep{peng2019transfer,Hahn2020}. We separate the detection of frequent and rare classes by introducing different training objectives. The special training objective for rare classes increases the proximity of the respective textual embeddings and the ontology embeddings of concepts. Our work also exploits data augmentation techniques, such as paraphrasing and guided text generation to aid sparse class detection and diversify the training data. 

We apply our methodology for the phenotype detection task with more than 15,000 concepts from the Human Phenotype Ontology (HPO)~\citep{kohler2017human} \footnote{In the medical text, the word ``phenotype'' refers to deviations from normal morphology, physiology, or behaviour, such as skin rash, hypoxemia, neoplasm, etc.~\citep{robinson2012deep}. Note the difference of the phenotypic information to the diagnosis information expressed in ICD-10 codes \citep{10665-42980}. These codes record patient health states mainly for billing purposes. The former contributes to the latter.}. The phenotyping task is an important Clinical NLP task that can improve the understanding of disease diagnosis~\citep{aerts2006gene,deisseroth2019clinphen,Liu2019,son2018deep,Xu2020}. It remains unexplored due to the complexity of the classification into that large amount of classes. We test our approach on clinical data, namely on electronic health records (EHRs) and radiology reports.

Our {\bf main contributions}: (1) Self-supervised methodology for contextual phenotype detection in clinical records. 
(2) Methodology for sparse class detection with the special training objective that increases proximity of contextual synonyms to ontology embeddings.
(3) Data augmentation methodology to further improve the detection of sparse classes.

Our self-supervised models improve the current SOTA on F1 up to 4.5 absolute points, while on Recall up to 4.0 absolute points for the phenotype detection task for clinical data, which demonstrates how relying on the context is essential for this type of data. Second, after fine-tuning, our model outperforms the fine-tuned BERT-based models with as little as 20\% of labelled data, which confirms efficiency of our self-supervised training objectives. Moreover, the extrinsic evaluation shows the benefits of using the phenotypes annotated by our model as features to predict ICU patient outcomes.

We present related work in Section~\ref{sec:related_work}, our phenotyping methods in Section~\ref{sec:methods}, and our experimental setup in Section~\ref{sec:experimental_setup}. Then, we present and discuss key results in Section~\ref{sec:results_and_discussion}. Finally, we conclude this work in Section~\ref{sec:conclusion}.

\section{Related Work}
\label{sec:related_work}
\begin{figure*}[!t]
 \centering
 \includegraphics[width=0.9\textwidth]{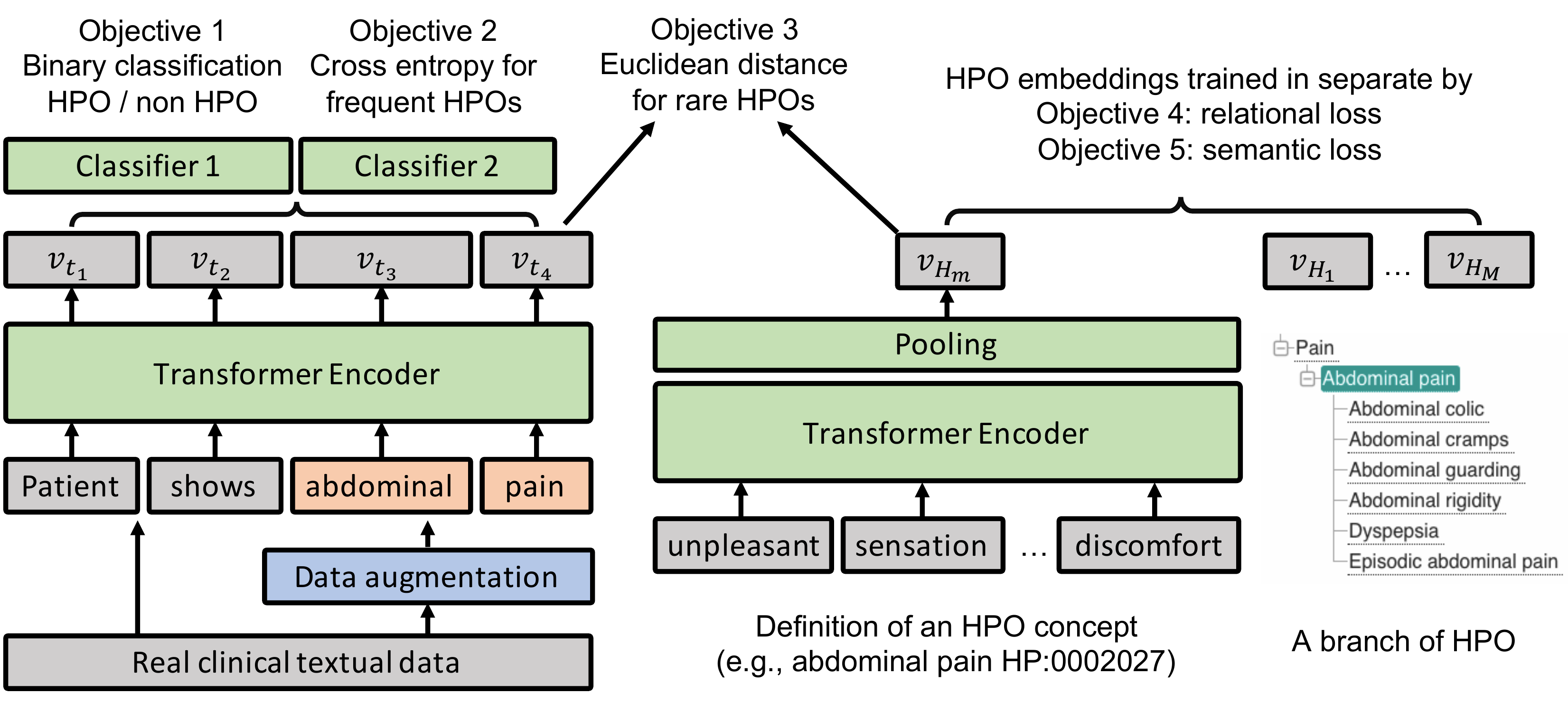}
 \caption{\textbf{Left}: Proposed model for phenotype annotation includes one Transformer encoder and two additional classifiers. The model is trained using the three self-supervised objectives (1, 2 and 3). \textbf{Middle and right}: Prior to the training of the phenotyping model, the other model for HPO embeddings is trained with relational and semantic losses (4, 5). Both Transformer encoders are initialised with ClinicalBERT but trained separately. The sub-figure of the HPO branch is taken from EMBL-EBI OLS \footnotemark. }\label{fig:architecture}
\end{figure*}

Most of the current methodologies for phenotype detection are supervised, BERT-based (e.g., BioBERT~\cite{biobert} or ClinicalBERT~\cite{alsentzer-etal-2019-publicly}) and dedicated to the detection of certain rather limited phenotypes or their groups \citep{liu-etal-2019-two,zhang2019unsupervised,Yang2020, franz2020deep, li2020behrt}. 

Unsupervised methods in the clinical NLP domain traditionally rely on the usage of ontologies and knowledge bases. Human Phenotype Ontology ~\cite{kohler2017human} is the most widely used ontology of phenotypes. 
The use of HPO in annotating phenotypic information automatically remains unexplored, mainly due to the complexity of formalising the task with over 15,000 concepts.

Such methods as MetaMap~\cite{Aronson2010} (the Mayo Clinic tool ~\cite{Shen2017} based on it), cTAKES~\cite{Savova2010}, NCBO~\cite{jonquet2009ncbo} and ClinPhen~\cite{deisseroth2019clinphen} follow similar pipelines and use linguistics patterns for shallow matching. 

More recently, unsupervised deep learning methods have been applied to the problem, which allowed to perform the semantic analysis and go beyond shallow matching~\citep{arbabi2019ncr,kraljevic2019medcat,Tiwari2020}. These approaches use non-contextualised embeddings, focus on the precision of detection with limited context exploitation.
For example, the authors in \citep{kraljevic2019medcat} propose a procedure to learn vectors of words enriched with their averaged context over the corpus to map them to correct medical concepts. 
We use contextualised word representations in contrast to all the related approaches and focus on recall.

\section{Methodology} 
\label{sec:methods}
This section introduces the problem of phenotype detection along with our self-supervised method. It elaborates our data augmentation strategies, selective supervision in low-resource conditions, and finally explains our inference algorithm.  

\footnotetext{EMBL-EBI OLS: \url{https://www.ebi.ac.uk/ols/ontologies/hp}}

\paragraph{Problem Definition} While annotating clinical text, clinicians usually relate HPOs to short spans, which usually have around 2-3 words depending on the corpus.~\footnote{This general observation is confirmed in our internal annotation procedure (see Section~\ref{ssec:sup_dataset})} Following this rationale, we define the phenotype annotation as a two-step process: (1) detect HPO-relevant text spans, and (2) assign respective HPO concepts to those spans. More formally, given a textual document $X = \{t_1, ..., t_N\}$ represented by a sequence of tokens, and a full set of HPO concepts $\mathcal{H} = \{H_1, ..., H_M\}$ under the root node \textit{Phenotypic Abnormality (HP:0000118)} (exclusive) of the HPO ontology, our goal is to model: (1) $p(1_H | t_n)$, which is the conditional probability of the token $t_n$ being HPO-relevant; (2) $p(H_m|t_n)$, which is the conditional probability if the HPO concept $H_m$ should be assigned to the token $t_n$.

In the self-supervised setting, we consider only the training examples with textual spans matched with exact match to the HPO concepts as defined by the ontology. The main assumption here is that by capturing context of those term spans, the model will be able to generalise and detect formally different HPO spans seen in the similar contexts as the HPO concepts (\aka contextual synonyms, for example,  \textit{Fever (HP:0001945)} will be matched to ``feverish''). To support this challenging setting, we have designed a series of relevant training objectives described below.

\paragraph{Training Objectives} As shown in Figure \ref{fig:architecture} (left), the proposed model for phenotype annotation consists of a Transformer encoder which is identical to and initialised with ClinicalBERT \citep{alsentzer-etal-2019-publicly}. Besides, there are two additional classifiers on the top of the Transformer encoder which predict if a token is HPO-relevant and assign HPO concepts to those HPO-relevant tokens.

We enrich the model with the following three training objectives. (1) \textbf{A binary cross-entropy loss} $\mathcal{L}_1$ to predict $p(1_H|t_n)$, where $1_H$ is 1 if $t_n$ is HPO-relevant, otherwise 0.
(2) \textbf{A cross-entropy loss} $\mathcal{L}_2$ with softmax to predict $p(H_m|t_n)$, which is defined over the most frequent HPO concepts found in the training data. The intuition behind this objective is to increase precision of prediction in the resulting performance.
(3) \textbf{The Euclidean distance} $\mathcal{L}_3$ between the token embedding $v_{t_n}$ and the respective HPO concept embedding $v_{H_m}$, which is defined to increase recall of the model and targets the detection of the rare HPO concepts. 

Note that the objectives above can be used for pre-training and further fine-tuning of the models in a way similar to BERT.

\paragraph{HPO Embeddings} Prior to the training of the phenotyping model, we build the knowledge graph (KG) embeddings for HPO concepts. Figure \ref{fig:architecture} (middle and right) shows that this KG model has a Transformer encoder which learns the embeddings of HPO concepts given their definitions. It is designed to encode both the hierarchical connections between HPO concepts and the semantics in definitions of HPO concepts, so that the similar HPO concepts have similar embeddings. Therefore, we consider two learning objectives. 

The first learning objective, namely \textbf{relational loss} $\mathcal{L}_4$, is to encourage the neighbouring HPO concepts to have similar embeddings and non-neighbouring HPO concepts to have different embeddings. The objective is implemented based on the distance of embeddings between neighbouring HPO concepts and non-neighbouring HPO concepts with softmax.

Second, the \textbf{semantic loss} $\mathcal{L}_5$ encourages the HPO embeddings to encode the semantics of input definitions and, more specifically, we adopt the skip-gram negative sampling \citep{mikolov:nips:2013}. 

\subsection{Data Augmentation}
\label{ssec:methods_data_augment}

There are two issues related to the creation of the training data by shallow matching: (1) this data can be too limited to help the model capture contextual phenotypes; (2) rare HPO concepts will not be found in the clinical text used for training and the model will not be able to detect them at the inference time. We are addressing those two problems by creating textual variants for existing HPO-relevant spans and generating context around rare HPO concepts.

\paragraph{HPO-relevant span variants with paraphrasing} are used to replace the original spans in the training sentences. We create the variants by using the standard lexical pivoting paraphrasing technique where equivalent phrases in one language are found by ``pivoting'' over a shared translation into another language~\cite{mallinson-etal-2017-paraphrasing}. We build the English-French-English pivot Seq2seq model. 

Phenotypes are also often inferred from ranges of numerical values. E.g., \textit{anemia (HP:0001903)} can be inferred from ``Hgb 5 g/dl''. We take the advantage of a series of reference laboratory values (from MIMIC~\cite{johnson2016mimic}) to create surrogates for the original names with numerical values. The named entities for which abnormal results are available are mapped to HPO concepts by an expert.

\paragraph{HPO context variants with synthetic text} are created with a Seq2Seq model, which is trained to generate the textual context conditioned on HPO-relevant spans. For example, the sentence ``patient was admitted with 
Angelman Syndrome to the ER'' is generated given the input ``Angelman Syndrome''.

\subsection{Decision Strategy for Inference}

At the phenotype annotation inference stage, we assume that the HPO-relevant spans of frequent HPO concepts can be detected by $p(1_H|t_n)$ and $p(H_m|t_n)$ with high precision, while the Euclidean distance between contextualised token embedding $v_{t_n}$ and HPO embedding $v_{H_m}$ should be able to find those of rare HPO concepts with good recall. More precisely, we formalise the decision strategy as Algorithm \ref{algo:decision_strategy_inference}.

\begin{algorithm}[h]

$X$ is the input sequence;

$\mathcal{H}$ stands for the full set of HPO concepts, $\mathcal{H}_{\text{freq}}\subset\mathcal{H}$ includes most frequent HPO concepts;

Initialise thresholds $\tau_p, \tau_d$ for $p(1_H|t_n)$ and distance function $D(v,u)$ respectively with pre-defined values;

\For{$t_n$ in $X = \{ t_1, t_2, \dots t_N \}$} {
    \If{$p(1_H|t_n) \geq \tau_p$}{
        $r_n = \arg\max_{H_m} p(H_m|t_n)$ where $H_m \in \mathcal{H}_{\text{freq}}$ ;
    }
    \ElseIf{$\min_{H_m} D(v_{t_n}, v_{H_m}) < \tau_d$}{
        $r_n = \arg\min_{H_m} D(v_{t_n}, v_{H_m})$;
    }
}
\Return{$\{ r_1, r_2, \dots, r_N \}$ }.
\caption{The decision strategy of inferring phenotype annotation.}
\label{algo:decision_strategy_inference}
\end{algorithm}

\section{Experimental Setup}
\label{sec:experimental_setup}
This section will introduce the datasets, implementation details, baselines and evaluation metrics.

\subsection{Pre-training Corpora}

\paragraph{EHR Corpus}

We use EHRs from the publicly available MIMIC-III database \citep{johnson2016mimic}.
Diseases of the circulatory system are the most common reasons for those ICU stays. We collect the training samples from 38,772 notes of brief hospital course in MIMIC-III's discharge summaries and 1.5M generated notes by using data augmentation which is also trained on MIMIC-III.

\paragraph{Scientific Literature Corpus} For the scientific text model, we use 119,924 PubMed abstracts \citep{cohan-etal-2018-discourse}, $\sim$ 180k lines from the Cochrane data \citep{ive:hal-01388655} and 1.5M generated notes by using data augmentation given PubMed abstracts. 

\paragraph{Ontologies} In the self-supervised setting, 
we consider HPO names, synonyms, abbreviations from the HPO as well as Unified Medical Language System (UMLS)~\citep{Bodenreider2004} with exact match in clinical text as training samples.

\subsection{Datasets}
\label{ssec:sup_dataset}

\begin{table}[!h]
\begin{center}
\renewcommand{\arraystretch}{1.1}
\scalebox{0.73}{
\begin{tabular}{|c|c|c|c|c|}
\hline
                        & MIMIC     & \covidi   & \covidii  & PubMed \\ \hline
\#, articles            & 242       & 67        & 100       & 228    \\ \hline       
avg \#, tokens          & 701.3     & 208.9     & 157.8     & 220.3  \\ \hline
avg \#, annotations     & 27.6      & 9.1       & 8.5       & 7.0    \\ \hline
avg \#, tok. / ann.     & 4.1       & 4.3       & 5.1       & 5.6    \\ \hline
avg HPO depth           & 4.4       & 4.4       & 4.4       & 4.8    \\ \hline
\#, unique HPO          & 946       & 91        & 201       & 422    \\ \hline
avg \# ann. / HPO       & 7.0       & 6.6       & 4.2       & 3.8    \\
\hline
\end{tabular}}
\end{center}
\caption{\label{table:data-stat} Statistics over the gold phenotype annotations of MIMIC, \covidi, \covidii, PubMed datasets. On average, each HPO appears less than 7.0 times.}
\end{table}

The following datasets are used as test data in the self-supervised setting, as well as train data in the supervised fine-tuning experiments.

\paragraph{Annotation Procedure}

To collect supervised datasets for evaluation and fine-tuning, we have annotated EHRs with HPO concepts with the help of three expert clinicians. The EHRs were pre-annotated with HPO concepts by keyword matching, and then the annotations were corrected by the three clinicians with consensus. The clinicians were specifically asked to identify \textbf{contextual synonyms} 
such as ``drop in blood pressure'' and ``BP of 79/48'' for \textit{Hypotension (HP:0002615)}.

\paragraph{MIMIC} We have created our own sub-corpus of 242 discharge summaries from MIMIC-III with gold annotations. We used 146 EHRs for fine-tuning in the low-resource setting. 48 and 48 EHRs are reserved respectively for validation and testing in both self-supervised and supervised settings.

\paragraph{COVID} We have collected and annotated two COVID datasets of short radiology reports: (1) \textbf{\covidi} has 67 radiology reports from the Italian Society of Medical and Interventional Radiology~\footnote{\url{https://www.sirm.org/category/senza-categoria/covid-19}} and (2) \textbf{\covidii} is the International dataset with 100 radiology reports presented by~\citep{cohen2020covidProspective}. From \covidii, we have selected the patients with the diagnosis of the COVID-19 viral pneumonia. We take all the unique patients and extracted the longest (in terms of the tokens count) records for those patients. Reports from both datasets often contain not only the findings, but also the brief patient history. Both datasets are used as test sets for the self-supervised model. In the experiments with supervision, \covidi was used to fine-tune and \covidii to test.

\paragraph{PubMed} 

To ensure comparison to the previous work, we also present our results for the PubMed dataset provided by \citep{groza2015automatic} which contains 228 abstracts annotated by the creators of HPO. 
The common HPOs in this dataset are neurodevelopmental and skeletal disorders (e.g. Angelman syndrome), which is a quite different group of phenotypes as compared to the groups represented in the MIMIC and COVID data. An important difference between our annotation procedure as described above and the human annotation for the PubMed data is that the latter instructed annotating HPO-relevant spans only if they were presented in a canonical form close to HPO names: for example, ``hypoplastic nails'' and ``nail hypoplasia'' were included, but not ``nails were hypoplastic''. We re-use the random split: 40 abstracts for training and 188 for testing following NCR's setting \citep{arbabi2019ncr}. The statistics over the dataset is in Table~\ref{table:data-stat}.

\subsection{Implementation Details \footnote{Due to the proprietary nature, the source code and gold annotations will not be shared publicly.}}
\label{ssec:implementation_details}

The Transformer encoders in Figure \ref{fig:architecture} are initialised by ClinicalBERT, the two classifiers are two dense layers and the pooling layer concatenates max and average pooling. The maximum input length is 64 tokens. The proposed models are pre-trained for 100k steps and fine-tuned for 5k steps with batch size 64. The set of frequent HPO concepts $\mid\mathcal{H}_{\text{freq}}\mid=400$ is decided by keyword matches.
For data augmentation, we train a Seq2Seq Transformer model on a range of parallel English-French corpora in the biomedical field, namely the European Medicines Agency, Corpus of Parallel Patent Applications and the PatTR corpora.\footnote{\url{http://statmt.org/wmt14/medical-task/}}
The Seq2Seq model is based on OpenNMT~\cite{klein-etal-2017-opennmt}. More details are given in Appendix \ref{appsec:training}.

\subsection{Setups}

In the self-supervised setting, we train our models using either EHRs corpus for MIMIC and COVID (E) or scientific literature corpus for PubMed (S). We experiment with two setups with and without data augmentation.

We also evaluate the efficiency of our training objectives for pre-training and fine-tune our models with all the available supervised data.

However, in the real-life clinical setting, human annotations are very costly thus particular attention should be paid to the learning efficiency with a very small amount of data. We simulate this low-resource scenario and analyse the annotation cost / performance benefit trade-offs for our model.
To be more precise, we run a set of experiments where each time we pick a certain percentage of training examples according to one of the following strategies: (1) \textbf{Random sampling}: the samples are selected at random; (2) \textbf{Uncertainty-based sampling}: the entropy score based on $p(H_m|t_n), m\in \{1,2,\dots,M\}$ is computed to measure the uncertainty of the self-supervised model for each sample, and then the samples with the highest uncertainty score are selected; (3) \textbf{Oracle}: we also count the number of mismatched phenotypes between the keyword-based and gold annotations, and the samples with the most mismatches are selected.

\subsection{Baselines}
\label{sec:baselines}

As baselines in the self-supervised setting, we report (1) Keyword: a naive method that simply matches HPO names, synonyms and abbreviations to text spans, (2) a range of text mining baselines (Clinphen \citep{deisseroth2019clinphen}, NCBO \citep{jonquet2009ncbo}, cTAKES \citep{Savova2010}, MetaMap \citep{Aronson2010}, MetaMapLite \citep{demner2017metamaplite}), and (3) two deep learning models (NCR \citep{arbabi2019ncr}, MedCAT \citep{kraljevic2019medcat}) which are trained without supervision.

In the selective supervision setting, we use pre-trained models and fine-tune them on the datasets. More specifically, we use (1) BERT-Base \citep{devlin-etal-2019-bert}, (2) BioBERT-Base v1.0 \citep{biobert} pre-trained on PubMed and PMC, (3) Clinical BERT \citep{alsentzer-etal-2019-publicly} pre-trained based on BioBERT and MIMIC-III discharge summaries, (4) SciBERT \citep{beltagy-etal-2019-scibert} pre-trained for scientific literature.

\begin{table*}[!htbp]
\begin{center}
\renewcommand{\arraystretch}{1.1}
\scalebox{0.8}{
\begin{tabular}{ |c|c|ccc|ccc| }
 \hline
 \multirow{2}{*}{Dataset} & \multirow{2}{*}{Method} & \multicolumn{3}{|c|}{Exact Match} & \multicolumn{3}{|c|}{Generalised Match} \\
    & & Precision  & Recall & F1 & Precision  &  Recall &  F1 \\
 \hline
 \multirow{5}{*}{MIMIC} & Keyword & 0.7496 & 0.5223 & 0.6156 & 0.7883 & 0.6587 & 0.7177 \\
 & NCR & 0.7747 & 0.4851 & 0.5967 & 0.8778 & 0.5917 & 0.7069 \\
 & NCBO & \textbf{0.9186} & 0.3901 & 0.5477 & \textbf{0.9632} & 0.4711 & 0.6328 \\
 & Ours (E)  & 0.7334 & \textbf{0.5619} & \textbf{0.6363} & 0.7706 & 0.6972 & 0.7320 \\
 & Ours (E) w. Augmented Data & 0.7235 & 0.5556 & 0.6285 & 0.7741 & \textbf{0.6997} & \textbf{0.7351} \\ 
 \hline
 
 \multirow{5}{*}{\covidi} & Keyword & 0.6897 & 0.4710 & 0.5597 & 0.6750 & 0.5579 & 0.6109 \\
 & NCR & 0.7873 & 0.4493 & 0.5721 & 0.8814 & 0.5481 & 0.6758 \\
 & NCBO & \textbf{0.8876} & 0.4293 & 0.5788 & \textbf{0.8857} & 0.5070 & 0.6449 \\
 & Ours (E) & 0.8617 & 0.4855 & 0.6211 & 0.8442 & 0.5657 & 0.6774 \\ 
 & Ours (E) w. Augmented Data & 0.8576 & \textbf{0.4909} & \textbf{0.6244} & 0.8800 & \textbf{0.5714} & \textbf{0.6929} \\
 \hline
 
 \multirow{5}{*}{\covidii} & Keyword & 0.8743 & 0.4514 & 0.5954 & 0.9268 & 0.5577 & 0.6963 \\
 & NCR & 0.7220 & 0.4703 & 0.5696 & 0.9136 & \textbf{0.6059} & \textbf{0.7286} \\
 & NCBO & \textbf{0.9006} & 0.4296 & 0.5817 & \textbf{0.9484} & 0.5128 & 0.6657 \\
 & Ours (E)  & 0.8517 & \textbf{0.4811} & \textbf{0.6149} & 0.9113 & 0.5814 & 0.7099 \\
 & Ours (E) w. Augmented Data & 0.8421 & 0.4757 & 0.6079 & 0.8859 & 0.5695 & 0.6933 \\
 \hline
 
 \multirow{5}{*}{PubMed} & Keyword & 0.7221 & 0.5277 & 0.6098 & 0.8735 & 0.7175 & 0.7879 \\
 & NCR & 0.7334 & \textbf{0.6443} & \textbf{0.6860} & 0.9131 & \textbf{0.8183} & \textbf{0.8631} \\
 & NCBO & \textbf{0.7948} & 0.4441 & 0.5698 & \textbf{0.9645} & 0.6227 & 0.7568 \\
 & Ours (S)  & 0.6756 & 0.5121 & 0.5826 & 0.8741 & 0.7035 & 0.7796 \\
 & Ours (S) w. Augmented Data & 0.6772 & 0.5627 & 0.6146 & 0.8818 & 0.7631 & 0.8182 \\
 \hline
\end{tabular}}
\end{center}
\caption{The proposed models in the self-supervised setting (without fine-tuning) achieved the best recall and F1 on MIMIC and COVID clinical text datasets. On PubMed which is scientific literature, our model clearly benefited from augmented data. Keyword, NCR and NCBO are reported as they achieve top F1 among the self-supervised baselines (Section \ref{sec:baselines}) and full results are reported in Appendix \ref{sec:appendix_self_supervised_res}. The notations ``Ours (E)'' and ``Ours (S)'' refer to the models pre-trained on the EHR corpus and the scientific literature corpus, respectively.}
\label{tab:selfsupervised_main}
\end{table*}
\begin{table*}[!h]
\begin{center}
\renewcommand{\arraystretch}{1.1}
\scalebox{0.8}{
\begin{tabular}{ |c|c|ccc|ccc| }
 \hline
 \multirow{2}{*}{Dataset} & \multirow{2}{*}{Method} & \multicolumn{3}{|c|}{Exact Match} & \multicolumn{3}{|c|}{Generalised Match} \\
    & & Precision  & Recall & F1 & Precision  &  Recall &  F1 \\
 \hline
 \multirow{2}{*}{MIMIC} 
 & Fine-tuned ClinicalBERT & 0.6962 & 0.5630 & 0.6225 & 0.8429 & 0.6980 & 0.7637  \\
 & Fine-tuned Ours (E) w. Augmented Data & \textbf{0.7141} & \textbf{0.7123} & \textbf{0.7132} & \textbf{0.8463} & \textbf{0.8380} & \textbf{0.8421}  \\
 \hline
 
 \multirow{2}{*}{\covidii} 
 & Fine-tuned ClinicalBERT & 0.6560 & 0.4138 & 0.5075 & 0.8063 & 0.5174 & 0.6303   \\ 
 & Fine-tuned Ours (E) w. Augmented Data &  \textbf{0.7027} & \textbf{0.6324} & \textbf{0.6657} & \textbf{0.8652} & \textbf{0.7980} & \textbf{0.8302}  \\
 \hline
 
 \multirow{2}{*}{PubMed} 
 & Fine-tuned ClinicalBERT & 0.5514 & 0.2449 & 0.3392 & 0.7715 & 0.4988 & 0.6059   \\
 & Fine-tuned Ours (S) w. Augmented Data & \textbf{0.7138} & \textbf{0.6618} & \textbf{0.6868} & \textbf{0.8959} & \textbf{0.8311} & \textbf{0.8623}   \\
 \hline
\end{tabular}}
\end{center}
\caption{The proposed model with fine-tuning in full achieved the best precision, recall and F1 scores on MIMIC, \covidii and PubMed. The \covidi is not reported as it is used to fine-tune the corresponding model. Only fine-tuned ClinicalBERT is reported as baseline because it achieves overall better F1 than fine-tuned BERT, BioBERT and SciBERT. Full results are available in Appendix \ref{sec:appendix_selective_supervision_res}. The notations ``Ours (E)'' and ``Ours (S)'' refer to the models pre-trained on the EHR corpus and the scientific literature corpus, respectively.}
\label{tab:supervised_main}
\end{table*}

\subsection{Metrics}
We report the scores of micro-averaged Precision, Recall and F1-score at the document level. Following the best practices and to make our work comparable with others, we adopt the evaluation strategy of ~\citep{Liu2019}. Thus, when we compute the following scores: (1) Exact match: only the exact same HPO annotations were counted as correct. (2) Generalised match: both the predicted and target HPO annotations are first extended to include all ancestors in HPO up until \textit{Phenotypic Abnormality (HP:0000118)} (exclusive). Then the HPO annotations are de-duplicated for each document and the scores are computed.

\section{Results and Discussion}
\label{sec:results_and_discussion}

This section discusses the results for the self-supervision and selective supervision settings.

\paragraph{Self-Supervised Setting} We report results of the self-supervised model for the MIMIC, COVID, and PubMed datasets in Table~\ref{tab:selfsupervised_main}. It compares the proposed model to the previous SOTA for the phenotyping task.
Our principal observation is that our method outperforms all the baselines in terms of F1 and recall across datasets for both the exact and generalised matches. 
For example, for the exact match, our best models obtain F1 gain of at least 0.02, 0.05, and 0.02 and Recall scores gain of at least 0.04, 0.02, and 0.01 for MIMIC, \covidi and \covidii, respectively. This confirms the efficiency of our methodology for the detection of contextual synonyms in clinical text. 

We note that our method does not give better performance for the PubMed dataset. 
We hypothesise that this happens due to the difference of gold annotation standards, as well as the fact that this dataset is oriented towards the detection of rare phenotypes with less frequent context patterns that are hence difficult to learn for our model.

\begin{figure*}[!t]
 \centering
 \includegraphics[width=1.0\textwidth]{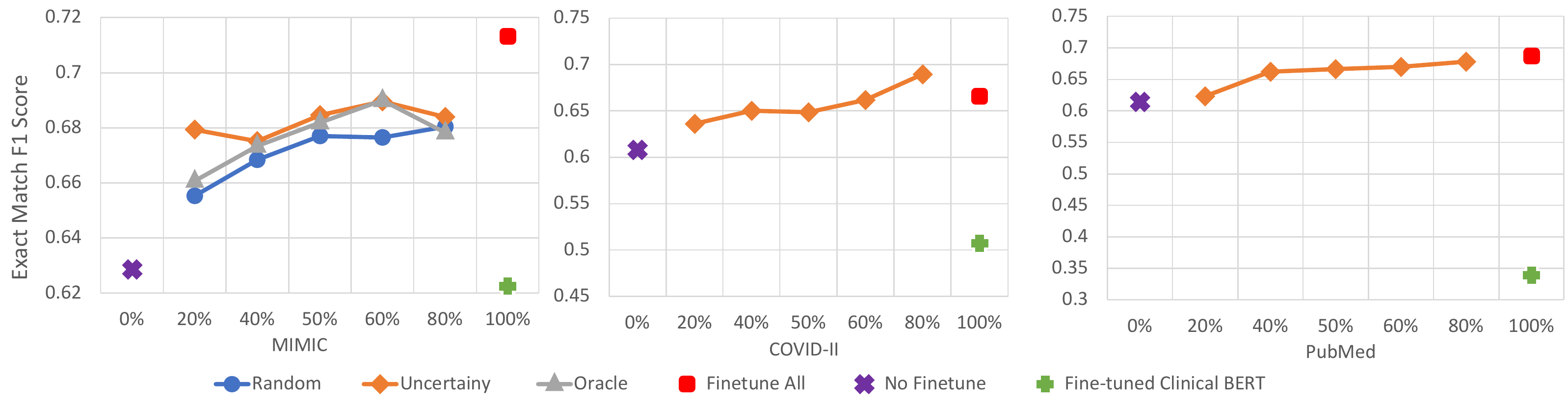}
 \caption{In the low resource setting with selective supervision, we pick subsets with 20\%, 40\%, 50\%, 60\%, 80\% labelled data to fine-tune. The uncertainty sampling strategy is consistently better than the other two strategies on MIMIC and then applied on \covidii and PubMed. The proposed models outperform fine-tuned BERT-based models with as little as 20\% of labelled data. Details in Table \ref{tab:selective_supervised_res} in Appendix~\ref{sec:appendix_selective_supervision_res}. Best to view in colours.}\label{fig:selective_supervision}
\end{figure*}
 
\begin{table*}[!h]
\begin{center}
\renewcommand{\arraystretch}{1.1}
\scalebox{0.8}{
\begin{tabular}{|c|c|c|ccc|}
\hline
\multirow{2}{*}{Task (Metric)} & \multirow{2}{*}{\shortstack{Structured \\ \cite{harutyunyan2019multitask}}}  & \multirow{2}{*}{Structured}  & \multicolumn{3}{|c|}{Structured + Phenotypes} \\ 
& & & + NCR  & + ClinicalBERT & + Ours \\ \hline
Length-of-stay (Kappa)  & 0.395 & 0.380 & 0.406 & 0.388 & \textbf{0.430} \\ \hline   
In-hospital Mortality (AUROC)  & 0.825 & 0.826 & 0.841 & 0.826 & \textbf{0.845} \\ \hline     
Decompensation (AUROC)  & 0.809 & 0.824 & 0.834 & 0.833 & \textbf{0.839} \\ \hline  
\end{tabular}}
\end{center}
\caption{Extrinsic evaluation on three ICU public benchmarks \cite{harutyunyan2019multitask} which are created based on MIMIC-III. The results of \cite{harutyunyan2019multitask} are reproduced by their code on the test set. }
\label{tab:secondary_eval}
\end{table*}

\paragraph{Low-Resource Setting} In this setting, we first study the efficiency of our self-supervised objectives for fine-tuning. Results are in Table \ref{tab:supervised_main} (more in Appendix~\ref{sec:appendix_selective_supervision_res}). Naturally fine-tuning leads to better automatic annotation accuracy on specific datasets. Our pre-training procedure is efficient and outperforms BERT-based models with at least 0.09, 0.16, 0.35 absolute increase in F1 (exact match) for the three datasets.
Our analysis of the annotation cost / performance benefit trade-offs demonstrated that with only 20\% of the training samples selected using the uncertainty criteria our fine-tuned model is able to achieve better F1 than ClinicalBERT which are fine-tuned on full training sets (see Figure \ref{fig:selective_supervision}). 

The HPOs are sparse (less than 7 annotations on average) in the datasets as shown in Table \ref{table:data-stat}. We further evaluate the model accuracy on annotating rare HPOs (any HPO excluding those from $\mathcal{H}_{\text{freq}}$ as defined in Section \ref{ssec:implementation_details}). Our fine-tuned model achieves F1 0.43 (exact match) and 0.60 (generalised match) on annotating rare HPOs while ClinicalBERT has F1 0.27 and 0.39 respectively and NCR achieves F1 0.34 and 0.47.

\paragraph{Qualitative Analysis} To get better insights into the model performance, we have manually eye balled outputs of our MIMIC and \covidi self-supervised model that achieves the best gain.

Our first observation is that our model is successful in capturing HPO-relevant contextual synonyms, which contributes to higher recall of our model. For example, ``low pressure'' and ``hypotensive'' are associated with \textit{Hypotension (HP:0002615)} and ``low platelets'' with \textit{Thrombocytopenia (HP:0001873)}.\footnote{All examples hereinafter are paraphrased.} Errors in the prediction mainly concern subtle distinctions between closely related phenotypes: \eg ``shortness of breath'' triggers prediction \textit{Respiratory Distress (HP:0002098)} whereas the gold label \textit{Dyspnea (HP:0002094)} is the generalisation of \textit{Respiratory Distress}.

For the more narrow-domain \covidi dataset, false negatives often concern missed radiographic concepts: \eg ``perihilar infiltration'' fails to trigger \textit{Pulmonary Infiltrates (HP:0002113)}. In distinction to above, errors of our selective supervision models are less coupled with radiographic observations, e.g., false negatives for \textit{Ankylosis (HP:0031013)}, \textit{Abnormal Ear Morphology (HP:0031703)} and \textit{Epileptic Spasm (HP:0011097)}.

\paragraph{Extrinsic Evaluation} We evaluate the benefit of using phenotypes extracted by our models as features to enhance performance on downstream tasks. Following the setting by \cite{harutyunyan2019multitask} with three public ICU benchmarks based on MIMIC-III, we train LSTMs with different input features: (1) 17 structured clinical features selected by \cite{harutyunyan2019multitask} like heart rate and temperature or (2) structured clinical features plus phenotypes annotated by NCR, ClinicalBERT and our fine-tuned model respectively. The patients with both structured clinical features and textual notes are collected, and as a result, there are 21,346 patients (25,106 admissions) for training (with 4-fold cross validation) and 3,824 patients (4,497 admissions) for testing. Table~\ref{tab:secondary_eval} shows that the LSTMs which are fed with structured clinical features and phenotypes annotated by our model are consistently better than others on all three benchmarks. This demonstrates that increasing recall in phenotyping is essential for downstream tasks.

\section{Conclusion}
\label{sec:conclusion}
In this paper, we have proposed a deep self-supervised phenotype annotation approach relying on contextualised word embeddings and data augmentation techniques. Our experimental results in a challenging sparse multi-class setting, with over 15,000 candidate HPO concepts, indicate that our methodology is particularly efficient to detect contextual mentions of phenotype concepts in clinical text. We demonstrate that increasing phenotyping recall is essential for downstream tasks.

\section{Ethics Considerations}

The study has been carried out in accordance with relevant guidelines and regulations for the MIMIC-III data. Other data used in this study can be accessed without any preliminary requests. Clinical experts received consulting fees for their work. The purpose of the developed models is to extract phenotypic information from unstructured healthcare data. This information is only to assist human medical experts in their decisions. Before the deployment in the actual clinical setting our methodology is subject to systematic debugging, extensive simulation, testing and validation under the supervision of expert clinicians. 

\section{Acknowledgement}

We would like to thank Dr. Garima Gupta, Dr. Deepa (M.R.S.H) and Dr. Ashok (M.S.) for helping us create gold-standard phenotype annotation data. We would also like to thank the four anonymous reviewers for the feedback.

\bibliographystyle{acl_natbib}
\bibliography{anthology,emnlp2021}

\clearpage
\onecolumn
\appendix
\section{Results of the Self-Supervised Models}
\label{sec:appendix_self_supervised_res}

\begin{table*}[!h]
\begin{center}
\renewcommand{\arraystretch}{1.1}
\scalebox{0.73}{
\begin{tabular}{ |c|ccc|ccc| }
 \hline
 \multicolumn{7}{|c|}{MIMIC} \\
 \hline
 Method & \multicolumn{3}{|c|}{Exact Match} & \multicolumn{3}{|c|}{Generalised Match} \\
     & Precision  & Recall & F1 & Precision  &  Recall &  F1 \\
 \hline
Keyword & 0.7496 & 0.5223 & 0.6156 & 0.7883 & 0.6587 & 0.7177 \\
NCR & 0.7747 & 0.4851 & 0.5967 & 0.8778 & 0.5917 & 0.7069 \\ 
Clinphen & 0.8147 & 0.3148 & 0.4541 & 0.9337 & 0.4066 & 0.5665 \\
NCBO & \textbf{0.9186} & 0.3901 & 0.5477 & \textbf{0.9632} & 0.4711 & 0.6328 \\ 
cTAKES & 0.8250 & 0.3259 & 0.4673 & 0.9321 & 0.4287 & 0.5872 \\ 
MetaMap & 0.7909 & 0.4062 & 0.5367 & 0.8835 & 0.5384 & 0.6691 \\
MetaMapLite & 0.7968 & 0.4358 & 0.5634 & 0.8766 & 0.5720 & 0.6923 \\
MedCAT (Medmentions) & 0.7290 & 0.3321 & 0.4563 & 0.8305 & 0.4711 & 0.6012 \\
MedCAT (UMLS) & 0.8630 & 0.3889 & 0.5362 & 0.9311 & 0.5231 & 0.6699 \\
\hline
Ours (E) & 0.7334 & \textbf{0.5619} & \textbf{0.6363} & 0.7706 & 0.6972 & 0.7320 \\
Ours (E) w. Augmented Data & 0.7235 & 0.5556 & 0.6285 & 0.7741 & \textbf{0.6997} & \textbf{0.7351} \\ 
\hline
\end{tabular}}
\end{center}
\caption{Results on MIMIC in the self-supervised setting.}
\label{tab:selfsupervised_mimic}
\end{table*}
\begin{table*}[!h]
\begin{center}
\renewcommand{\arraystretch}{1.1}
\scalebox{0.73}{
\begin{tabular}{ |c|ccc|ccc| }
 \hline
 \multicolumn{7}{|c|}{\covidi} \\
 \hline
 Method & \multicolumn{3}{|c|}{Exact Match} & \multicolumn{3}{|c|}{Generalised Match} \\
     & Precision  & Recall & F1 & Precision  &  Recall &  F1 \\
 \hline
Keyword & 0.6897 & 0.4710 & 0.5597 & 0.6750 & 0.5579 & 0.6109 \\
NCR & 0.7873 & 0.4493 & 0.5721 & 0.8814 & 0.5481 & 0.6758 \\
Clinphen & 0.8259 & 0.3351 & 0.4768 & 0.8693 & 0.4042 & 0.5518 \\
NCBO & \textbf{0.8876} & 0.4293 & 0.5788 & \textbf{0.8857} & 0.5070 & 0.6449 \\
cTAKES & 0.7305 & 0.1866 & 0.2973 & 0.8285 & 0.3112 & 0.4524 \\
MetaMap & 0.8023 & 0.3750 & 0.5111 & 0.8990 & 0.5039 & 0.6458 \\
MetaMapLite & 0.7765 & 0.3587 & 0.4907 & 0.8992 & 0.4914 & 0.6355 \\
MedCAT (Medmentions) & 0.7519 & 0.3514 & 0.4790 & 0.8284 & 0.4940 & 0.6189 \\
MedCAT (UMLS) & 0.6293 & 0.2645 & 0.3724 & 0.8295 & 0.4171 & 0.5551 \\
\hline
Ours (E) & 0.8617 & 0.4855 & 0.6211 & 0.8442 & 0.5657 & 0.6774 \\ 
Ours (E) w. Augmented Data & 0.8576 & \textbf{0.4909} & \textbf{0.6244} & 0.8800 & \textbf{0.5714} & \textbf{0.6929} \\
\hline
\multicolumn{7}{|c|}{\covidii} \\
 \hline
 Method & \multicolumn{3}{|c|}{Exact Match} & \multicolumn{3}{|c|}{Generalised Match} \\
     & Precision  & Recall & F1 & Precision  &  Recall &  F1 \\
 \hline
Keyword & 0.8743 & 0.4514 & 0.5954 & 0.9268 & 0.5577 & 0.6963 \\
NCR & 0.7220 & 0.4703 & 0.5696 & 0.9136 & \textbf{0.6059} & \textbf{0.7286} \\
Clinphen & 0.7789 & 0.3290 & 0.4626 & 0.9038 & 0.4256 & 0.5787 \\
NCBO & \textbf{0.9006} & 0.4296 & 0.5817 & \textbf{0.9484} & 0.5128 & 0.6657 \\
cTAKES & 0.7684 & 0.2098 & 0.3296 & 0.9158 & 0.3327 & 0.4881 \\
MetaMap & 0.8517 & 0.3218 & 0.4672 & 0.9437 & 0.4152 & 0.5767 \\
MetaMapLite & 0.7828 & 0.3261 & 0.4604 & 0.9494 & 0.4431 & 0.6042 \\
MedCAT (Medmentions) & 0.7599 & 0.3046 & 0.4349 & 0.8757 & 0.4284 & 0.5753 \\
MedCAT (UMLS) & 0.8333 & 0.2586 & 0.3947 & 0.9368 & 0.3675 & 0.5280 \\
\hline
Ours (E) & 0.8517 & \textbf{0.4811} & \textbf{0.6149} & 0.9113 & 0.5814 & 0.7099 \\
Ours (E) w. Augmented Data & 0.8421 & 0.4757 & 0.6079 & 0.8859 & 0.5695 & 0.6933 \\
\hline
\end{tabular}}
\end{center}
\caption{Results on \covidi and \covidii in the self-supervised setting.}\label{tab:selfsupervised_covid}
\end{table*}
\begin{table*}[!h]
\begin{center}
\renewcommand{\arraystretch}{1.1}
\scalebox{0.73}{
\begin{tabular}{ |c|ccc|ccc| }
 \hline
 \multicolumn{7}{|c|}{PubMed} \\
 \hline
 Method & \multicolumn{3}{|c|}{Exact Match} & \multicolumn{3}{|c|}{Generalised Match} \\
     & Precision  & Recall & F1 & Precision  &  Recall &  F1 \\
 \hline
Keyword & 0.7221 & 0.5277 & 0.6098 & 0.8735 & 0.7175 & 0.7879 \\
NCR & 0.7334 & \textbf{0.6443} & \textbf{0.6860} & 0.9131 & \textbf{0.8183} & \textbf{0.8631} \\
Clinphen & 0.6352 & 0.3926 & 0.4853 & 0.9240 & 0.5095 & 0.6568 \\
NCBO & \textbf{0.7948} & 0.4441 & 0.5698 & \textbf{0.9645} & 0.6227 & 0.7568 \\
cTAKES & 0.5602 & 0.2216 & 0.3175 & 0.8953 & 0.3479 & 0.5011 \\
MetaMap & 0.7167 & 0.4966 & 0.5867 & 0.9076 & 0.6671 & 0.7690 \\
MetaMapLite & 0.7057 & 0.4334 & 0.5370 & 0.8978 & 0.5934 & 0.7146 \\ 
MedCAT (Medmentions) & 0.5362 & 0.2089 & 0.3007 & 0.7387 & 0.3066 & 0.4333 \\
MedCAT (UMLS) & 0.7636 & 0.4237 & 0.5450 & 0.9376 & 0.5903 & 0.7245 \\
\hline
Ours (S)  & 0.6756 & 0.5121 & 0.5826 & 0.8741 & 0.7035 & 0.7796 \\
Ours (S) w. Augmented Data & 0.6772 & 0.5627 & 0.6146 & 0.8818 & 0.7631 & 0.8182 \\
\hline
\end{tabular}}
\end{center}
\caption{Results on PubMed in the self-supervised setting.}
\label{tab:selfsupervised_pubmed}
\end{table*}

\clearpage

\section{Selective Supervision Results in Low-Resource Setting}
\label{sec:appendix_selective_supervision_res}

\begin{table*}[!h]
\begin{center}
\renewcommand{\arraystretch}{1.1}
\scalebox{0.73}{
\begin{tabular}{ |c|ccc|ccc| }
 \hline
 \multicolumn{7}{|c|}{MIMIC} \\
 \hline
 Method & \multicolumn{3}{|c|}{Exact Match} & \multicolumn{3}{|c|}{Generalised Match} \\
     & Precision  & Recall & F1 & Precision  &  Recall &  F1 \\
 \hline
BERT            &   0.7132 & 0.5617 & 0.6285 & 0.8434 & 0.6844 & 0.7557  \\
BioBERT         &   0.6864 & 0.5728 & 0.6245 & 0.8335 & 0.7021 & 0.7622  \\
ClinicalBERT   &   0.6962 & 0.5630 & 0.6225 & 0.8429 & 0.6980 & 0.7637  \\
SciBERT         &   0.6898 & 0.5407 & 0.6062 & 0.8269 & 0.6671 & 0.7385  \\ \hline
Ours (E) w. Augmented Data  &   \textbf{0.7141} & \textbf{0.7123} & \textbf{0.7132} & \textbf{0.8463} & \textbf{0.8380} & \textbf{0.8421}  \\
\hline
\multicolumn{7}{|c|}{\covidii} \\
\hline
 Method & \multicolumn{3}{|c|}{Exact Match} & \multicolumn{3}{|c|}{Generalised Match} \\
     & Precision  & Recall & F1 & Precision  &  Recall &  F1 \\
 \hline
BERT            &    0.6144 & 0.3549 & 0.4499 & 0.8193 & 0.4760 & 0.6022   \\
BioBERT         &    0.5858 & 0.3922 & 0.4699 & 0.7781 & 0.5201 & 0.6235   \\
ClinicalBERT   &    0.5711 & 0.4095 & 0.4770 & 0.7680 & 0.5039 & 0.6085   \\
SciBERT         &    0.6560 & 0.4138 & 0.5075 & 0.8063 & 0.5174 & 0.6303   \\ \hline
Ours (E) w. Augmented Data  &    \textbf{0.7027} & \textbf{0.6324} & \textbf{0.6657} & \textbf{0.8652} & \textbf{0.7980} & \textbf{0.8302}  \\
\hline
\multicolumn{7}{|c|}{PubMed} \\
\hline
 Method & \multicolumn{3}{|c|}{Exact Match} & \multicolumn{3}{|c|}{Generalised Match} \\
     & Precision  & Recall & F1 & Precision  &  Recall &  F1 \\
 \hline
BERT            &    0.5103 & 0.2400 & 0.3265 & 0.7530 & 0.4795 & 0.5859   \\
BioBERT         &    0.4828 & 0.2459 & 0.3258 & 0.7716 & 0.4911 & 0.6002   \\
ClinicalBERT   &    0.5514 & 0.2449 & 0.3392 & 0.7715 & 0.4988 & 0.6059   \\
SciBERT         &    0.4967 & 0.2177 & 0.3027 & 0.7187 & 0.4638 & 0.5638   \\ \hline
Ours (E) w. Augmented Data  &   \textbf{0.7138} & \textbf{0.6618} & \textbf{0.6868} & \textbf{0.8959} & \textbf{0.8311} & \textbf{0.8623}   \\
\hline
\end{tabular}}
\end{center}
\caption{Results on MIMIC, \covidii and PubMed with supervision. All models are fine-tuned with full training samples.}
\label{tab:supervised}
\end{table*}
\begin{table*}[!h]
\begin{center}
\renewcommand{\arraystretch}{1.1}
\scalebox{0.73}{
\begin{tabular}{ |c|c|ccc|ccc| }
 \hline
 \multicolumn{8}{|c|}{MIMIC} \\
 \hline
 Sampling Strategy & Ratio & \multicolumn{3}{|c|}{Exact Match} & \multicolumn{3}{|c|}{Generalised Match} \\
     & & Precision  & Recall & F1 & Precision  &  Recall &  F1 \\
 \hline
\multirow{5}{*}{Random} & 20\%    &   0.6496 & 0.6609 & 0.6552 & 0.7760 & 0.8074 & 0.7914  \\
& 40\% & 0.6441 & 0.6943 & 0.6683 & 0.7709 & 0.8408 & 0.8044 \\
& 50\% & 0.6529 & 0.7030 & 0.6770 & 0.7755 & 0.8357 & 0.8045 \\
& 60\% & 0.6457 & 0.7104 & 0.6765 & 0.7744 & 0.8367 & 0.8044 \\
& 80\% & 0.6508 & 0.7129 & 0.6804 & 0.7765 & 0.8419 & 0.8078 \\ \hline
\multirow{5}{*}{Oracle} & 20\%    & 0.6344 & 0.6894 & 0.6607 & 0.7521 & 0.8200 & 0.7846 \\
& 40\% & 0.6349 & 0.7166 & 0.6733 & 0.7631 & 0.8453 & 0.8021 \\
& 50\% & 0.6422 & 0.7265 & 0.6818 & 0.7562 & 0.8575 & 0.8037 \\
& 60\% & 0.6530 & 0.7314 & 0.6900 & 0.7626 & 0.8582 & 0.8076 \\
& 80\% & 0.6370 & 0.7252 & 0.6782 & 0.7707 & 0.8524 & 0.8095 \\ \hline
\multirow{5}{*}{Uncertainty} & 20\% &   0.6707 & 0.6881 & 0.6793 & 0.7667 & 0.8142 & 0.7898 \\
& 40\% & 0.6592 & 0.6918 & 0.6751 & 0.7692 & 0.8269 & 0.7970 \\
& 50\% & 0.6503 & 0.7228 & 0.6846 & 0.7659 & 0.8476 & 0.8047 \\
& 60\% & 0.6580 & 0.7240 & 0.6895 & 0.7737 & 0.8541 & 0.8119 \\
& 80\% & 0.6499 & 0.7215 & 0.6839 & 0.7722 & 0.8514 & 0.8099 \\ \hline
\multicolumn{8}{|c|}{\covidii} \\
\hline
 Sampling Strategy & Ratio & \multicolumn{3}{|c|}{Exact Match} & \multicolumn{3}{|c|}{Generalised Match} \\
     & & Precision  & Recall & F1 & Precision  &  Recall &  F1 \\
 \hline
\multirow{5}{*}{Uncertainty} & 20\% &   0.6990 & 0.5838 & 0.6362 & 0.8680 & 0.7261 & 0.7907 \\
& 40\% & 0.7129 & 0.5973 & 0.6500 & 0.8796 & 0.7303 & 0.7980 \\
& 50\% & 0.6978 & 0.6054 & 0.6483 & 0.8719 & 0.7610 & 0.8127 \\
& 60\% & 0.7220 & 0.6108 & 0.6618 & 0.8874 & 0.7603 & 0.8190 \\
& 80\% & 0.7573 & 0.6324 & 0.6892 & 0.8951 & 0.7750 & 0.8307 \\ \hline
\multicolumn{8}{|c|}{PubMed} \\
\hline
 Sampling Strategy & Ratio & \multicolumn{3}{|c|}{Exact Match} & \multicolumn{3}{|c|}{Generalised Match} \\
     & & Precision  & Recall & F1 & Precision  &  Recall &  F1 \\
 \hline
\multirow{5}{*}{Uncertainty} & 20\% &   0.6899 & 0.5685 & 0.6233 & 0.8852 & 0.7766 & 0.8273 \\
& 40\% & 0.7156 & 0.6161 & 0.6621 & 0.8922 & 0.7949 & 0.8408 \\
& 50\% & 0.6877 & 0.6463 & 0.6663 & 0.8798 & 0.8226 & 0.8502 \\
& 60\% & 0.7014 & 0.6414 & 0.6701 & 0.8883 & 0.8141 & 0.8496 \\
& 80\% & 0.7028 & 0.6550 & 0.6781 & 0.8867 & 0.8192 & 0.8516 \\ \hline
\end{tabular}}
\end{center}
\caption{Results on MIMIC, \covidii and PubMed in the selective supervision setting. The F1 scores of exact match correspond to Figure \ref{fig:selective_supervision}.}
\label{tab:selective_supervised_res}
\end{table*}

\clearpage

\section{Implementation Details} 
\label{appsec:training}

The data processing and model are developed by Python 3.6. Besides our own code, we use open-sourced third-party libraries including Matplotlib \citep{thomas_a_caswell_2020_4268928}, Numpy \citep{harris2020array}, Pandas, Pronto, Scikit-learn \citep{scikit-learn}, Transformers \citep{wolf-etal-2020-transformers}, Tensorboard, Pytorch \citep{NEURIPS2019_bdbca288} (v1.7, CUDA 10.1), Tqdm, Xmltodict. The number of learnable parameters is close to a BERT-base model.  On two NVIDIA TITANX GPUs, it takes around 24 hours to pre-train and 1.2 hours to fine-tune.

\begin{table}[!h]
	\centering
	\renewcommand{\arraystretch}{1.1}
	\begin{tabular}{|c|c|} 
	    \hline
	    \multicolumn{2}{|c|}{Self-supervised training} \\ \hline
		Optimiser               & AdamW \\
		Training steps          & 100k    \\
		Learning rate           & 1e-4  \\
		Batch size              & 64    \\
		Vocab size              & 28996 \\
		Maximum input length    & 64    \\ \hline
		\multicolumn{2}{|c|}{Fine-tuning} \\ \hline
		Optimiser               & AdamW \\
		Training steps          & 5k    \\
		Learning rate           & 1e-4  \\
		Batch size              & 64    \\
		Vocab size              & 28996 \\
		Maximum input length    & 64    \\
		\hline
		\multicolumn{2}{|c|}{HPO Embeddings} \\ \hline
		Optimiser               & AdamW \\
		Training steps          & 30k    \\
		Learning rate           & 2e-5  \\
		Batch size              & 64    \\
		Vocab size              & 28996 \\
		Maximum input length    & 64    \\
		\hline
	\end{tabular}
	\caption{Hyper-parameters for training are decided empirically on the validation set.}
	\label{tab:hyperparam-erac}
\end{table}

\end{document}